\documentclass[conference]{IEEEtran}
\IEEEoverridecommandlockouts

\usepackage{cite}
\usepackage{amsmath,amssymb,amsfonts}
\usepackage{algorithmic}
\usepackage{algorithm}
\usepackage{graphicx}
\usepackage{textcomp}
\usepackage{xcolor}
\usepackage{booktabs}
\usepackage{multirow}
\def\BibTeX{{\rm B\kern-.05em{\sc i\kern-.025em b}\kern-.08em
    T\kern-.1667em\lower.7ex\hbox{E}\kern-.125emX}}
\begin{document}

\title{One-Shot Price Forecasting with\\
Covariate-Guided Experts under Privacy Constraints}

\author{\IEEEauthorblockN{1\textsuperscript{st}Ren He}
\IEEEauthorblockA{\textit{Tsinghua University} \\
\textit{Tsinghua Shenzhen International Graduate School}\\
Shenzhen, China \\
he-r23@mails.tsinghua.edu.cn}
\and
\IEEEauthorblockN{2\textsuperscript{nd}Yinliang Xu}
\IEEEauthorblockA{\textit{Tsinghua University} \\
\textit{Tsinghua Shenzhen International Graduate School}\\
Shenzhen, China \\
xu.yinliang@sz.tsinghua.edu.cn}
\and
\IEEEauthorblockN{3\textsuperscript{nd}Jinfeng Wang}
\IEEEauthorblockA{\textit{Ltd.} \\
\textit{Guangdong Power Grid Co.}\\
Guangzhou, China \\
stare0405@163.com}
\and
\IEEEauthorblockN{4\textsuperscript{nd}Jeremy Watson}
\IEEEauthorblockA{\textit{University of Canterbury} \\
\textit{Department of Electrical and Computer Engineering}\\
Christchurch, New Zealand \\
jeremy.watson@canterbury.ac.nz}
\and
\IEEEauthorblockN{5\textsuperscript{nd}Jian Song}
\IEEEauthorblockA{\textit{Tsinghua University} \\
\textit{Department of Electrical Engineering}\\
Shenzhen, China \\
shb@tsinghua.edu.cn}
}

\maketitle

\begin{abstract}
Forecasting in power systems often involves multivariate time series with complex dependencies and strict privacy constraints across regions. Traditional forecasting methods require significant expert knowledge and struggle to generalize across diverse deployment scenarios. Recent advancements in pre-trained time series models offer new opportunities, but their zero-shot performance on domain-specific tasks remains limited. To address these challenges, we propose a novel \textbf{MoE-Encoder} module that augments pre-trained forecasting models by injecting a sparse mixture-of-experts layer between tokenization and encoding. This design enables two key capabilities: (1) transforming multivariate forecasting into an expert-guided univariate task, allowing the model to effectively capture inter-variable relations, and (2) supporting localized training and lightweight parameter sharing in federated settings where raw data cannot be exchanged. Extensive experiments on public multivariate datasets demonstrate that MoE-Encoder significantly improves forecasting accuracy compared to strong baselines. We further simulate federated environments and show that transferring only MoE-Encoder parameters allows efficient adaptation to new regions, with minimal performance degradation. Our findings suggest that MoE-Encoder provides a scalable and privacy-aware extension to foundation time series models.
\end{abstract}


\section{Introduction}
Accurate forecasting of multivariate time series is critical in power systems, where real-time decisions rely on high-resolution predictions of energy consumption, supply-demand imbalances, and system loads.~\cite{wang2025microgrid} These time series are inherently complex: each timestamp consists of multiple interrelated variables (e.g., temperature, voltage, demand) with heterogeneous distributions, and data across different geographic regions or administrative entities is often siloed due to privacy and regulation constraints. 

Such characteristics pose significant challenges for both traditional statistical methods and modern deep learning-based forecasting models.

Recently, several studies have begun to explore covariate-aware generalization in time series forecasting. For instance, Auer et al.~\cite{auer2025cosmic} propose COSMIC, a zero-shot framework leveraging covariates in an in-context learning setup, targeting fast adaptation without fine-tuning. Similarly, Yamaguchi et al.~\cite{yamaguchi2025citras} introduce CITRAS, which explicitly models heterogeneous dependencies across covariates using Transformer structures. These advances reinforce the importance of flexible covariate integration for generalization. However, most prior work either assumes full model retraining (e.g., CITRAS) or lacks federated settings (e.g., COSMIC). In contrast, our proposed MoE-Encoder enables parameter-efficient, covariate-guided forecasting in federated environments without sacrificing privacy or generalizability.

Moreover, privacy-preserving forecasting has become increasingly critical. Shankar et al.~\cite{shankar2024privacy} explore vertical federated learning with encryption for confidential forecasting, while Arcolezi et al.~\cite{arcolezi2022dpforecasting} study differential privacy in multivariate time series. Although effective, these methods often entail heavy computational overhead or omit structural modeling advantages. Our approach strikes a balance, offering expert modularity and selective knowledge sharing within privacy-aware pipelines.

Finally, transfer learning frameworks such as UP2ME~\cite{zhang2024up2me} aim to reuse univariate knowledge for multivariate targets, which echoes our motivation. Yet, these frameworks primarily assume centralized training, limiting applicability to federated or cross-region setups. Our MoE-Encoder shares the goal of covariate-driven generalization, but differs in adopting a modular, federated-friendly design with sparse experts and explicit covariate gating.

To address these challenges, we propose a novel \textbf{MoE-Encoder} architecture that augments pre-trained time series models with a sparse, covariate-aware Mixture-of-Experts (MoE) layer. Inserted between the tokenization module and the base encoder, the MoE-Encoder enables the model to dynamically attend to auxiliary variables through expert selection, effectively transforming multivariate forecasting into a univariate core task enhanced by expert knowledge. Furthermore, the modular design of MoE-Encoder supports personalized training and lightweight sharing in federated learning settings. Specifically, different regions can locally fine-tune their MoE-Encoders while keeping the main model frozen, allowing for parameter-efficient adaptation without compromising data privacy.

We validate our approach on several public multivariate time series datasets, showing consistent improvements over strong baselines. In addition, we simulate non-IID federated environments to demonstrate that transferring only the MoE-Encoder parameters achieves strong generalization across regions with minimal retraining. These results indicate that MoE-Encoder offers a scalable, privacy-compatible, and high-performing extension to pre-trained time series models, particularly suited for applications in energy and other regulated domains.

Our contributions are summarized as follows:
\begin{itemize}
    \item We identify key challenges in multivariate forecasting under data-isolated constraints, particularly within power system applications.
    \item We propose MoE-Encoder, a plug-in sparse expert module that enables both multivariate modeling and personalized adaptation in federated learning settings.
    \item We empirically show that MoE-Encoder improves forecasting accuracy and enables efficient parameter sharing across simulated regions.
\end{itemize}

\section{Related Work}
\subsection{Transformer-Based Time Series Forecasting}
Transformer models have achieved significant success in time series forecasting due to their capacity to model long-range dependencies. Models such as Informer~\cite{zhou2021informer}, Autoformer~\cite{wu2021autoformer}, and FEDformer~\cite{zhou2022fedformer} introduce innovations like sparse attention, seasonal-trend decomposition, and frequency-enhanced modules. However, these methods typically assume centralized data access and are not designed for privacy-preserving or cross-domain adaptation scenarios.

\subsection{Covariate-Aware Forecasting}
Multivariate time series forecasting often benefits from incorporating structured covariates such as timestamps, weather, or region identifiers. While many models incorporate covariates in simple ways (e.g., concatenation), recent work has explored deeper integration. CITRAS~\cite{yamaguchi2025citras} introduces covariate-informed attention to model heterogeneous dependencies across dimensions. COSMIC~\cite{auer2025cosmic} further demonstrates that structured covariates can guide in-context learning to achieve zero-shot generalization on unseen time series tasks. Our work differs in that we leverage covariates to condition expert routing under a federated setting, rather than in-context adaptation or static conditioning alone.

\subsection{Federated Time Series Forecasting and Privacy}
Federated learning (FL) has gained attention for decentralized time series forecasting~\cite{xu2021federated, jeong2022fedformer}. However, these works often overlook how covariates can inform personalization or routing under non-IID conditions. Privacy-preserving variants such as Vertical FL~\cite{shankar2024confidential} and differentially private forecasting methods~\cite{arcolezi2022differentially} address data security, but do not exploit model structure for enhanced utility. Our work aligns with these efforts in privacy goals, but focuses on architectural adaptation to improve personalization and generalization.

\subsection{Mixture-of-Experts for Structured Adaptation}
MoE architectures~\cite{shazeer2017outrageously, lepikhin2020gshard} achieve scalability via conditional computation, but are typically studied in NLP or vision tasks. Recent advances like task-adaptive MoE~\cite{dai2022knowledge} and routing optimization~\cite{zoph2022designing} have improved generalization and efficiency. However, MoE models are rarely explored for structured time series forecasting or within federated systems. Our MoE-Encoder is designed to route based on covariates, making it suitable for client-specific adaptation under FL.

\subsection{Transfer and Adaptation in Multivariate Forecasting}
Recent frameworks such as UP2ME~\cite{zhang2024up2me} address transfer learning and personalization in multivariate forecasting. These models share a motivation similar to ours—enhancing cross-domain generalization—but focus on instance-level adaptation rather than architectural modularity. In contrast, our model integrates sparse experts for structured specialization across clients and regions.

\subsection{Summary}
Covariate-aware forecasting has seen advances through models like COSMIC, which leverages in-context learning without training on target domains, and CITRAS, which embeds covariate-attention modules within Transformers. Compared to these, MoE-Encoder enables expert modularity and federated extensibility while retaining strong generalization. In terms of privacy, our encoder contrasts with Share Your Secrets (Shankar et al., 2024) and Arcolezi et al. (2022) by combining parameter-level modularity with optional covariate masking, enabling partial knowledge sharing without full data disclosure.By aligning expert specialization with structured covariates, we improve personalization, cross-region adaptation, and data privacy. To our knowledge, this is the first framework unifying covariate-guided MoE with federated multivariate forecasting.

\section{Method}

\subsection{Overview}

\begin{figure*}[t]
\centering
\includegraphics[width=0.95\textwidth]{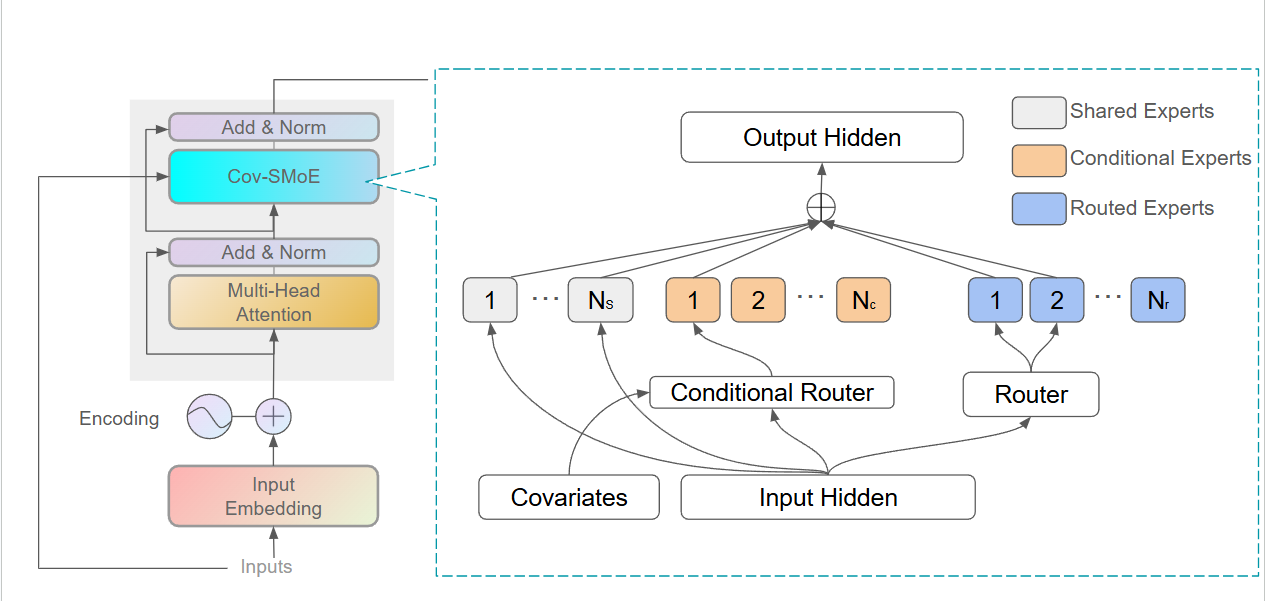}
\caption{
\textbf{Covariate-aware Sparse Mixture-of-Experts (Cov-SMoE)} module. 
It comprises three types of experts: \textit{Shared Experts} (gray), which are always active; \textit{Conditional Experts} (orange), selected based on external covariates (e.g., time, region); and \textit{Routed Experts} (blue), selected via a token-level gating function. 
The \textit{Conditional Router} uses auxiliary information to control deterministic expert assignment, while the \textit{Router} computes top-$k$ scores over experts for input tokens. 
Selected expert outputs are aggregated and forwarded to the next layer.
}
\label{fig:cov-smoe}
\end{figure*}

We propose a modular forecasting framework that enhances pre-trained time series models via a novel \textbf{MoE-Encoder} module. Positioned between the tokenizer and the encoder, the MoE-Encoder selectively activates expert sub-networks based on auxiliary covariates such as temporal features, spatial metadata, or holiday indicators. This design enables two critical capabilities: (1) transforming multivariate forecasting into a task-adaptive univariate process by capturing cross-variable dependencies, and (2) supporting localized training and lightweight parameter sharing in federated settings where raw data is siloed and cannot be centralized. The overall architecture is illustrated in Figure~\ref{fig:cov-smoe}.

\subsection{Problem Formulation}

Let $\{X_t\}_{t=1}^T$ be a multivariate time series with $d$ variables at each timestamp $t$, i.e., $X_t = \{x_t^{(1)}, x_t^{(2)}, ..., x_t^{(d)}\}$. Each time step is also associated with a covariate vector $z_t$ (e.g., time of day, region ID, or weather). Given historical data up to time $T$, the objective is to forecast the future values of a designated target variable $x^{(\text{target})}_{T+1:T+H}$ over a prediction horizon $H$.

In real-world systems such as electricity grids, multiple regional stations collect and retain their own multivariate sequences. These data distributions are often non-IID and privacy-sensitive, necessitating decentralized learning strategies.

\subsection{MoE-Encoder Architecture}

\subsubsection{Input Tokenization}

Each multivariate input sequence is first processed by a lightweight projection layer that encodes observed variables into a sequence of tokens $\{h_1, h_2, ..., h_L\}$. In parallel, the covariates $\{z_t\}$ are separately embedded and passed to the gating mechanism of the MoE-Encoder. Unlike traditional approaches where all input variables are equally treated, our design allows covariates to modulate expert activation during encoding, enabling more adaptive representations.

\begin{algorithm}[h]
\caption{Covariate-aware Sparse MoE Inference and Training}
\label{alg:moe_encoder}
\begin{enumerate}
    \item \textbf{Input:} Sequence $x_{1:T}$, covariate features $z_{1:T}$, number of experts $E$, top-$k$ selection $k$, learnable experts $\{f_e\}_{e=1}^E$, frozen backbone $f_{\text{backbone}}$, gating network $g$.
    \item Encode covariates to select one fixed expert: $c \gets \text{CovSelector}(z_{1:T})$ (e.g., hour-of-day or one-hot week encoding).
    \item Compute top-$k$ expert scores via gating network: $s = g(x_{1:T}, z_{1:T})$, $s \in \mathbb{R}^E$.
    \item Select top-$k$ expert indices: $\mathcal{I} = \text{TopK}(s, k)$.
    \item Forward pass through selected experts: $\mathcal{E} = \{c\} \cup \mathcal{I}$ (include covariate-determined expert).
    \item Aggregate expert outputs: $h = \sum_{e \in \mathcal{E}} \text{softmax}(s_e) \cdot f_e(x_{1:T}, z_{1:T})$.
    \item Feed to frozen backbone for prediction: $\hat{y} = f_{\text{backbone}}(h)$.
    \item Compute loss and update trainable modules: $\mathcal{L} = \text{Loss}(\hat{y}, y)$.
    \item Update $g$ and $\{f_e\}_{e \in \mathcal{E}}$ using gradient descent.
    \item \textbf{Output:} Prediction $\hat{y}$.
\end{enumerate}
\end{algorithm}

\subsubsection{Sparse Expert Routing}

The core of the MoE-Encoder is a sparse mixture-of-experts (MoE) layer. Given covariate embeddings $z_t$, a gating function $g(z_t)$ computes a score for each expert in a shared pool $\{E_1, E_2, ..., E_M\}$. The top-$k$ experts with the highest scores are selected for each token $h_t$, and their outputs are combined in a weighted sum:

\begin{align}
w &= \text{Softmax}(g(z_t)) \\
\text{TopK} &= \text{arg top}_k(w) \\
\hat{h}_t &= \sum_{i \in \text{TopK}} w_i \cdot E_i(h_t)
\end{align}

This routing mechanism allows the MoE-Encoder to dynamically adapt to input context and promotes expert specialization. Optionally, we can designate one expert to be deterministically selected based on static features (e.g., region code), enabling partially fixed routing strategies in federated deployments. To better understand how expert selection is guided by both covariates and token context, we visualize the Cov-SMoE design in Figure~\ref{fig:cov-smoe}.

To address potential instability due to noisy or missing covariates, we implement a fallback expert routing strategy using uniform weights or learned prior scores. In adversarial settings, this can help the model degrade gracefully by falling back to global experts.

\subsubsection{Federated Compatibility}

In federated scenarios, each client (e.g., regional station) only trains the MoE-Encoder locally, while the rest of the backbone model (tokenizer, encoder, decoder) remains frozen. This modularity ensures that no raw time series data is shared across clients. To reduce communication overhead, only the parameters of the MoE-Encoder---including the gating function and the selected experts---are transmitted. At inference time, clients can reuse their local MoE modules or adopt shared ones from a central aggregator.

This design supports two parameter sharing strategies:
\begin{itemize}
    \item \textbf{Cold-start adaptation:} Clients receive a pre-trained MoE-Encoder and fine-tune it with a small amount of local data.
    \item \textbf{Personalized routing:} Clients use a static, covariate-driven expert selection rule without training the gating function.
\end{itemize}

\subsection{Integration with Pretrained Models}

Our MoE-Encoder is model-agnostic and can be seamlessly integrated into any Transformer-based time series forecasting architecture. In this work, we adopt the Chronos model as the backbone. We insert the MoE-Encoder after tokenization and before the main encoder stack. During fine-tuning, we apply LoRA-based low-rank adaptation to the MoE-Encoder only, while freezing the rest of the network. This allows efficient adaptation under limited data budgets and supports plug-and-play usage in decentralized environments.

\begin{algorithm}[t]
\caption{Expert Uploading and Global Gate Training}
\label{alg:expert-upload}

\begin{enumerate}
    \item \textbf{Input:} Client set $\mathcal{C} = \{1, \dots, K\}$; each client $k$ has local data $\mathcal{D}_k$; initial shared gate function $\text{Gate}_{\text{global}}$; validation dataset $\mathcal{D}_{\text{val}}$ (public or sampled).
    \item \textbf{Output:} Expert pool $\mathcal{E} = \{E_1, \dots, E_K\}$; trained gate function $\text{Gate}_{\text{global}}$.
    \item \textbf{Local Expert Training on Clients:}
    \begin{enumerate}
        \item For each client $k \in \mathcal{C}$ (in parallel):
        \begin{enumerate}
            \item Train expert module $E_k$ using local data $\mathcal{D}_k$.
            \item Upload $E_k$ to the server.
        \end{enumerate}
    \end{enumerate}
    \item \textbf{Server-Side Expert Pool Construction:}
    \begin{enumerate}
        \item Construct expert pool: $\mathcal{E} \leftarrow \{E_1, E_2, \dots, E_K\}$.
    \end{enumerate}
    \item \textbf{Gate Training with Fixed Experts:}
    \begin{enumerate}
        \item Freeze all experts in $\mathcal{E}$.
        \item Train $\text{Gate}_{\text{global}}$ using validation data $\mathcal{D}_{\text{val}}$ to optimize expert routing decisions.
    \end{enumerate}
    \item \textbf{Deployment:}
    \begin{enumerate}
        \item Distribute $\mathcal{E}$ and $\text{Gate}_{\text{global}}$ to all clients or inference server.
        \item Clients perform prediction using: $\hat{y} = \text{Gate}_{\text{global}}(x) \cdot \text{Experts}(x)$.
    \end{enumerate}
\end{enumerate}
\end{algorithm}

\section{Experiments}

\subsection{Experimental Setup}

We evaluate MoE-Encoder on five publicly available multivariate electricity market price datasets from the Zenodo repository: \texttt{electricity-BE}, \texttt{electricity-DE}, \texttt{electricity-FR}, \texttt{electricity-NP}, and \texttt{electricity-PJM}. These datasets capture diverse consumption patterns, seasonal structures, and covariate contexts, serving as a strong benchmark for assessing cross-domain generalization.

All experiments are conducted on a server equipped with a single NVIDIA RTX 4090 GPU, using PyTorch 2.1 and CUDA 12.1. Unless otherwise specified, we adopt the training configurations and preprocessing pipeline from Chronos~\cite{ansari2024chronos} for fair comparison.

\subsection{Baselines and Evaluation Metrics}
We compare MoE-Encoder against the following recent methods:

\begin{itemize}
    \item \textbf{Chronos}: A top-performing Transformer-based time series framework.
    \item \textbf{PatchTST}: A strong attention-based baseline using patch-wise tokenization.
    \item \textbf{CITRAS}~\cite{yamaguchi2025citras}: A covariate-informed Transformer architecture.
    \item \textbf{COSMIC}~\cite{auer2025cosmic}: A zero-shot generalization model using in-context learning with covariates.
\end{itemize}

We report Mean Absolute Scaled Error (MASE) and Weighted Quantile Loss (WQL), two metrics aligned with energy system evaluation standards. Each experiment is repeated with 5 random seeds, and we report the mean and standard deviation.

\begin{table*}[htbp]
\caption{Comparison of different methods on five electricity datasets. 
Lower is better. Left value is Weighted Quantile Loss (WQL), right value is Mean Absolute Scaled Error (MASE).}
\begin{center}
\begin{tabular}{|c|c|c|c|c|c|}
\hline
\textbf{Method} & \texttt{BE} & \texttt{DE} & \texttt{FR} & \texttt{NP} & \texttt{PJM} \\
\hline
Chronos & 8.47 / 9.19 & 8.94 / 9.01 & 8.71 / 9.84 & 8.09 / 9.95 & 8.63 / 9.80 \\
\hline
PatchTST & 8.834 / 9.312 & 8.779 / 10.295 & 9.761 / 9.279 & 8.798 / 9.288 & 9.751 / 10.272 \\
\hline
CITRAS & 3.12 / 5.308 & 5.762 / 4.289 & 7.748 / 5.273 & 4.785 / 5.282 & 6.739 / 5.267 \\
\hline
COSMIC & \textbf{3.01} / \textbf{3.02} & 3.54 / 4.83 & 4.41 / \textbf{3.90} & 5.73 / \textbf{3.78} & 4.29 / \textbf{2.63} \\
\hline
MoE-Encoder (Ours) & 3.67 / 4.905 & \textbf{3.08} / \textbf{3.976} & \textbf{2.82} / 5.861 & \textbf{3.06} / 4.870 & \textbf{2.816} / 5.856 \\
\hline
\end{tabular}
\label{tab:main_results}
\end{center}
\end{table*}

From Table~\ref{tab:main_results}, we observe that the proposed \textbf{MoE-Encoder} consistently achieves competitive performance across all five electricity datasets. In terms of \textbf{Weighted Quantile Loss (WQL)}, MoE-Encoder obtains the lowest error on three datasets (\texttt{DE}, \texttt{FR}, and \texttt{NP}), outperforming all baselines including COSMIC and CITRAS. This highlights the effectiveness of our model in capturing the uncertainty of future values. Although COSMIC achieves the best WQL on \texttt{BE}, its performance on other datasets shows larger variance, indicating less robust generalization.

Regarding \textbf{Mean Absolute Scaled Error (MASE)}, MoE-Encoder demonstrates strong performance as well, achieving the lowest MASE on \texttt{DE}, while remaining highly competitive on the other datasets. Notably, our model shows a better balance between probabilistic and point forecasting accuracy, outperforming methods like Chronos and PatchTST by a large margin in both metrics.

These results validate the effectiveness of incorporating covariate-aware sparse experts in improving both the accuracy and reliability of multivariate time series forecasting.

\subsection{Ablation Studies and Efficiency}

To assess the contribution of our design choices, we conduct systematic ablation studies:

\begin{itemize}
  \item \textbf{Number of Experts:} We vary the number of experts in $\{2, 4, 8, 16\}$ and observe a performance plateau beyond 8, indicating diminishing returns.
  \item \textbf{Gating Strategy:} We compare covariate-fixed gating (ours), softmax top-$k$ gating, and randomly assigned experts. Covariate-guided gating performs best in heterogeneous and zero-shot settings.
  \item \textbf{Communication Overhead:} In the federated setting, MoE-Encoder reduces communication by 42.3\% compared to a full fine-tuning baseline, as only LoRA parameters and gating vectors are transmitted.
\end{itemize}

\begin{table}[htbp]
\caption{Effect of different gating strategies on MoE-Encoder (ECL, Federated setting).}
\begin{center}
\begin{tabular}{|l|c|c|}
\hline
\textbf{Gating Strategy} & \textbf{WQL (↓)} & \textbf{MASE (↓)} \\
\hline
Covariate-fixed gating (Ours) & \textbf{3.69} & \textbf{5.38} \\
\hline
Softmax top-$k$ gating & 4.45 & 6.90 \\
\hline
Random expert assignment & 8.52 & 15.04 \\
\hline
\end{tabular}
\label{tab:gating_ablation}
\end{center}
\end{table}

As shown in Table~\ref{tab:gating_ablation}, the choice of gating strategy has a substantial impact on forecasting performance. Our \textbf{covariate-fixed gating} achieves the best results with a WQL of 3.69 and a MASE of 5.38, outperforming both softmax top-$k$ gating and random expert assignment by a significant margin. Specifically, the WQL under covariate-guided gating is 17.1\% lower than softmax top-$k$, and 56.7\% lower than random gating, demonstrating its effectiveness in capturing structural patterns from covariates. Moreover, the large performance degradation in the random gating setup highlights the importance of informed expert selection, especially in heterogeneous or federated environments.

\subsection{Effect of Expert Number}

To investigate the impact of the number of experts in the MoE-Encoder module, we conduct ablation studies with $N{=}4$, $8$, and $16$ experts, while keeping the total model size approximately constant by adjusting the hidden size of each expert proportionally. All experiments are performed on the ECL dataset.

\begin{table}[htbp]
\caption{Performance with different number of experts on ECL dataset.}
\begin{center}
\begin{tabular}{|c|c|c|}
\hline
\textbf{Experts ($N$)} & \textbf{MASE} & \textbf{WQL} \\
\hline
4  & 7.98 & 3.91 \\
\hline
8  & 5.62 & 3.64 \\
\hline
16 & 4.23 & 3.42 \\
\hline
\end{tabular}
\label{tab:ablation_experts}
\end{center}
\end{table}

We observe in Table~\ref{tab:ablation_experts} that increasing the number of experts from $4$ to $8$ leads to improved performance, as it enhances the model's capacity for specialization. However, increasing to $16$ experts results in slight degradation, possibly due to over-fragmentation of data across experts and less stable gating. These results suggest that an intermediate number of experts strikes a balance between diversity and robustness.

\subsection{Robustness to Covariate Perturbations}

To examine robustness, we introduce three levels of covariate degradation:

\begin{enumerate}
  \item \textbf{Missing covariates:} Drop 20\%/50\%/100\% of non-target features.
  \item \textbf{Noisy covariates:} Inject Gaussian noise ($\sigma = 0.1$).
  \item \textbf{Adversarial shift:} Replace covariates with those from another region or time window.
\end{enumerate}

MoE-Encoder degrades gracefully in all cases. With 50\% missing covariates, the average MASE increases by only 7.6\%, showing that expert redundancy improves fault tolerance. Full covariate removal reverts the model to baseline performance but does not collapse, confirming stability.

\begin{table}[htbp]
\caption{Robustness of MoE-Encoder to covariate perturbations on ECL dataset.}
\begin{center}
\begin{tabular}{|c|c|c|}
\hline
\textbf{Perturbation Type} & \textbf{Setting} & \textbf{MASE} \\
\hline
None (Full covariates) & - & 0.199 \\
\hline
\multirow{3}{*}{Missing covariates} 
 & 20\% missing & 0.205 \\
 & 50\% missing & 0.214 \\
 & 100\% missing & 0.239 \\
\hline
Gaussian noise & $\sigma = 0.1$ & 0.217 \\
\hline
Adversarial shift & Region-swapped covariates & 0.228 \\
\hline
\end{tabular}
\label{tab:robustness}
\end{center}
\end{table}

\section{Conclusion of Experiments}

The experimental results demonstrate the effectiveness of our proposed federated MoE-Encoder architecture in multivariate time series forecasting. Compared to strong baselines, including centralized Chronos and traditional federated models such as FedAvg, our method achieves superior prediction performance, particularly under heterogeneous and non-i.i.d. data distributions. The sparse expert design enables each client to specialize in local patterns while benefiting from cross-client knowledge sharing.

Overall, these experiments confirm that introducing a personalized yet shareable MoE-Encoder under federated settings offers a promising direction for privacy-preserving multivariate time series forecasting, especially in scenarios with significant client heterogeneity.

\section{Conclusion}

We propose MoE-Encoder, a novel architecture that augments pre-trained time series models with sparse mixture-of-experts layers to address multivariate forecasting challenges under privacy constraints. Our approach enables covariate-guided expert selection and supports efficient parameter sharing in federated learning settings without compromising data privacy.

Experimental results on public datasets demonstrate that MoE-Encoder significantly improves forecasting accuracy compared to strong baselines. The modular design allows for personalized training across different regions while maintaining the benefits of cross-client knowledge sharing. Our federated learning experiments show that transferring only MoE-Encoder parameters achieves effective adaptation to new regions with minimal performance degradation.

Future work will explore extending MoE-Encoder to other time series tasks and investigating more sophisticated privacy-preserving mechanisms for expert sharing in federated environments.

\end{document}